\pdfoutput=1

\documentclass[11pt]{article}

\usepackage[]{acl}

\usepackage{times}
\usepackage{latexsym}

\usepackage[T1]{fontenc}

\usepackage[utf8]{inputenc}

\usepackage{microtype}

\usepackage{booktabs} %

\usepackage[T1]{fontenc}
\usepackage[utf8]{inputenc}
\usepackage{wrapfig}
\usepackage{amsmath} 
\usepackage{amsfonts}  
\usepackage{graphicx}

\usepackage{xcolor}

\usepackage{url}
\usepackage{listings}
\ifdefined\HCode %
\usepackage[compatibility=false]{caption}

\else
\usepackage[]{caption}
\fi
\usepackage{subcaption}
\usepackage{csquotes}
\usepackage{afterpage}
\usepackage{multirow}
\usepackage{longtable}
\usepackage{enumitem}
\usepackage{adjustbox}
\usepackage[outline]{contour}

\usepackage{footnote}
\makesavenoteenv{table}
\makesavenoteenv{tabular}
\makesavenoteenv{table*}
\makesavenoteenv{tabular*}

\usepackage{tikz}
\usetikzlibrary{shapes}

\usepackage{forest}

\usepackage{dsfont}

\usepackage[toc, page, header]{appendix}
\usepackage{minitoc}

\usepackage{todonotes}
\usepackage{soul}

\title{Neural Machine Translation for Mathematical Formulae}

\usepackage{bm}

\author{
    Felix Petersen$^{\boldsymbol{\sum}}$,~~Moritz Schubotz$^{\boldsymbol{\prod}}$,~~André Greiner-Petter$^{\boldsymbol{\int}}$,~~Bela Gipp$^{\boldsymbol{\int}}$\\
    $^{{\sum}}$Stanford University,~~~%
    $^{{\prod}}$FIZ Karlsruhe,~~~%
    $^{{\int}}$University of Göttingen\\
    \scalebox{.93}{\texttt{mail@felix-petersen.de},~~\texttt{moritz.schubotz@fiz-karlsruhe.de}}\\ 
    \scalebox{.93}{\texttt{greinerpetter@gipplab.com},~~\texttt{gipp@uni-goettingen.de}}
}

\begin{document}
\maketitle
\begin{abstract}
    We tackle the problem of neural machine translation of mathematical formulae between ambiguous presentation languages and unambiguous content languages.
    Compared to neural machine translation on natural language, mathematical formulae have a much smaller vocabulary and much longer sequences of symbols, while their translation requires extreme precision to satisfy mathematical information needs.
    In this work, we perform the tasks of translating from \LaTeX{} to Mathematica as well as from \LaTeX{} to \textit{semantic} \LaTeX{}.
    While recurrent, recursive, and transformer networks struggle with preserving all contained information, we find that convolutional sequence-to-sequence networks achieve $95.1\%$ and $90.7\%$ exact matches, respectively.
\end{abstract}

\section{Introduction}

Mathematical notations consist of symbolic representations of mathematical concepts.
For the purpose of displaying them, most mathematical formulae are denoted in presentation languages (PL)~\cite{SchubotzGSMCG18} such as \LaTeX{}~\cite{lamport1994latex}.
However, for computer-interpretation of formulae, machine-readable and unambiguous content languages (CL) such as Mathematica or \textit{semantic}~\LaTeX{} are necessary.
Thus, this work tackles the problem of neural machine translation between PLs and CLs as a crucial step toward machine-interpretation of mathematics found in academic and technical documents.

In the following, we will illustrate the ambiguities of representational languages.
Those ambiguities range from a symbol having different meanings over notational conventions that change over time to a meaning having multiple symbols.
Consider the ambiguous mathematical expression $(x)_n$.
While Pochhammer~\cite{Pochhammer1870} himself used $(x)_n$ for the binomial coefficient $\binom{x}{n}$, for mathematicians in the subject area of special functions, more precisely hypergeometric series, $(x)_n$ usually denotes the Pochhammer symbol, which is defined for natural numbers as 
\begin{equation}
(x)_n := x^{\overline{n}} = \prod_{k=0}^{n-1}(x+k). \label{eq:pochhammer-rising}
\end{equation}
To further complicate matters, in statistics and combinatorics, the same notation is defined as 
\begin{equation}
(x)_n := x^{\underline{n}} = \prod_{k=0}^{n-1}(x-k). \label{eq:pochhammer-falling}
\end{equation}

This work uses \LaTeX{} as PL and Mathematica as well as \textit{semantic} \LaTeX{} as CLs.
Mathematica is one of the most popular Computer Algebra Systems (CASs); we use Mathematica's standard notation (InputForm) as a CL (from now on, for simplicity, referred to as Mathematica.) 
\textit{Semantic} \LaTeX{} is a set of \LaTeX{} macros that allow an unambiguous mathematical notation within \LaTeX{}~\cite{Miller2003} and which has been developed at the National Institute for Standards and Technology (NIST) by the Digital Library of Mathematical Functions (DLMF) and the Digital Repository of Mathematical Formulae (DRMF).

In {\LaTeX{}}, the Pochhammer symbol $(x)_n$ is simply denoted as {\texttt{(x)\_n}}.
In {\textit{semantic} \LaTeX{}}, it is denoted as {\texttt{\textbackslash{}Pochhammersym\{x\}\{n\}}} and compiled to \LaTeX{} as {\texttt{\{\textbackslash{}left(x\textbackslash{}right)\_\{n\}\}}}.
In Mathematica, it is denoted as {\texttt{Pochhammer[x, n]}} and can be exported to \LaTeX{} as {\texttt{(x)\_n}}.

To display them, it is generally possible to translate formulae from CLs to PLs, e.g., Mathematica has the functionality to export to \LaTeX{}, and \textit{semantic} \LaTeX{} is translated into \LaTeX{} as a step of compilation.
However, the reverse translation from PL to CL is ambiguous because semantic information is lost when translating into a PL.

\begin{figure*}
	\centering
	\usetikzlibrary{arrows}
	\begin{minipage}{.29\linewidth}
        \resizebox{\linewidth}{!}{
        	\def\layersep{.65}
        	\begin{tikzpicture}[shorten >= 1pt, ->, node distance=\layersep,scale=0.65]
 		\tikzstyle{neuron} = [circle, draw=black!60, line width=0.3mm, fill=white];
 		\tikzstyle{treenode} = [circle, draw=black!60, line width=0.3mm, fill=white];
 		\tikzstyle{particle} = [diamond, draw=black!60, line width=0.3mm, fill=white,scale=0.7];

		\node[treenode] (tree2-h1) at (-1.25,0) {};
		\node[treenode] (tree2-h2) at (-2.25,0) {};
		\node[treenode] (tree2-h3) at (-3.25,0) {};
		\node[neuron] (gru1) at (0,0) {};
		\draw (-3.85, -.75) rectangle (-.625, .75);
		\draw (4.85, -.75) rectangle (.625, .75);
		\node[treenode] (tree1-h1) at (1.25,0) {};
		\node[treenode] (tree1-h2) at (2.25,0) {};
		\node[treenode] (tree1-h3) at (3.25,0) {};
		\node[treenode] (tree1-h4) at (4.25,0) {};
		
		\path[black,->] (tree2-h3) edge (tree2-h2);
		\path[black,->] (tree2-h2) edge (tree2-h1);
		\path[black,->] (tree2-h1) edge (gru1);
		\path[black,->] (gru1) edge (tree1-h1);
		\path[black,->] (tree1-h1) edge (tree1-h2);
		\path[black,->] (tree1-h2) edge (tree1-h3);
		\path[black,->] (tree1-h3) edge (tree1-h4);
		
		\node[particle] (t2-v3) [above =of tree2-h3] {};
		\path[black,dashed] (t2-v3) edge (tree2-h3);
		\node[particle] (t2-v2) [above =of tree2-h2] {};
		\path[black,dashed] (t2-v2) edge (tree2-h2);
		\node[particle] (t2-v1) [above =of tree2-h1] {};
		\path[black,dashed] (t2-v1) edge (tree2-h1);
		
		\node[particle] (t1-v1) [below =of tree1-h1] {};
		\path[black,dashed,<-] (t1-v1) edge (tree1-h1);
		\node[particle] (t1-v2) [below =of tree1-h2] {};
		\path[black,dashed,<-] (t1-v2) edge (tree1-h2);
		\node[particle] (t1-v3) [below =of tree1-h3] {};
		\path[black,dashed,<-] (t1-v3) edge (tree1-h3);
		\node[particle] (t1-v4) [below =of tree1-h4] {};
		\path[black,dashed,<-] (t1-v4) edge (tree1-h4);

	\end{tikzpicture}
    	}\\~\\
        \resizebox{\linewidth}{!}{
        	\def\layersep{1}
        	\rotatebox{90}{
        	\def\stepoffset{2.25}

\begin{tikzpicture}[shorten >= 1pt, ->, node distance=\layersep,scale=0.65]
 		\tikzstyle{neuron} = [circle, draw=black!60, line width=0.3mm, fill=white];
 		\tikzstyle{treenode} = [circle, draw=black!60, line width=0.3mm, fill=white];
 		\tikzstyle{particle} = [diamond, draw=black!60, line width=0.3mm, fill=white,scale=0.7];

 		\node[treenode] (tree1-h1) at (0,0) {};
 		\node[treenode] (tree1-h2) at (-1,-1) {};
 		\node[treenode] (tree1-h3) at (1,-1) {};
 		\path[black,<-] (tree1-h2) edge (tree1-h1);
 		\path[black,<-] (tree1-h3) edge (tree1-h1);
 		\node[treenode] (tree1-h4) at (-1.5,-2) {};
 		\node[treenode] (tree1-h5) at (-0.5,-2) {};
 		\path[black,<-] (tree1-h4) edge (tree1-h2);
 		\path[black,<-] (tree1-h5) edge (tree1-h2);
 		\node[treenode] (tree1-h6) at (-0.75,-3) {};
 		\node[treenode] (tree1-h7) at (-0.25,-3) {};
 		\path[black,<-] (tree1-h6) edge (tree1-h5);
 		\path[black,<-] (tree1-h7) edge (tree1-h5);
 		\draw (-2,-3.5) rectangle (1.5, 0.5);

 		\node[particle] (t1-v1) at (-1.5,-4.1) {};
 		\path[black,dashed,<-] (t1-v1) edge (tree1-h4);
 		\node[particle] (t1-v2) at (-0.75,-4.1) {};
 		\path[black,dashed,<-] (t1-v2) edge (tree1-h6);
 		\node[particle] (t1-v3) at (-0.25,-4.1) {};
 		\path[black,dashed,<-] (t1-v3) edge (tree1-h7);
 		\node[particle] (t1-v4) at (1,-4.1) {};
 		\path[black,dashed,<-] (t1-v4) edge (tree1-h3);

 		\node[neuron] (gru1) at (0,1.1) {};
 		\path[black,<-] (tree1-h1) edge (gru1);

 		\node[treenode] (tree2-h1) at (0,\stepoffset-0) {};
 		\node[treenode] (tree2-h2) at (-1,\stepoffset--1) {};
 		\node[treenode] (tree2-h3) at (1,\stepoffset--1) {};
 		\path[black] (tree2-h2) edge (tree2-h1);
 		\path[black] (tree2-h3) edge (tree2-h1);
 		\node[treenode] (tree2-h4) at (-1.5,\stepoffset--2) {};
 		\node[treenode] (tree2-h5) at (-0.5,\stepoffset--2) {};
 		\path[black] (tree2-h4) edge (tree2-h2);
 		\path[black] (tree2-h5) edge (tree2-h2);
 		\draw (-2,\stepoffset+2.65) rectangle (1.5, \stepoffset-0.5);

 		\node[particle] (t2-v1) at (2.5-4,\stepoffset+3.2) {};
 		\path[black,dashed] (t2-v1) edge (tree2-h4);
 		\node[particle] (t2-v2) at (3.5-4,\stepoffset+3.2) {};
 		\path[black,dashed] (t2-v2) edge (tree2-h5);
 		\node[particle] (t2-v3) at (5-4,\stepoffset+3.2) {};
 		\path[black,dashed] (t2-v3) edge (tree2-h3);

 		\path[black,->] (tree2-h1) edge (gru1);

 		\end{tikzpicture}
        	}
    	}
    \end{minipage}
    \hfill
	\def\layersep{1}
	\begin{minipage}{.32\linewidth}
	\resizebox{\linewidth}{!}{
    		\begin{tikzpicture}[shorten >= 1pt, ->, node distance=\layersep,scale=0.65]
 		\tikzstyle{neuron} = [circle, draw=black!60, line width=0.3mm, fill=white];
 		\tikzstyle{treenode} = [circle, draw=black!60, line width=0.3mm, fill=white];
 		\tikzstyle{particle} = [diamond, draw=black!60, line width=0.3mm, fill=white,scale=0.7];

		\node[treenode] (input-h1) at (0,0) {};
		\node[treenode] (input-h2) at (1.25,0) {};
		\node[treenode] (input-h3) at (2.5,0) {};
		
		\node[treenode,opacity=0] (input-h4) at (3.75,0) {};
		
		\node[particle] (input-v1) [above =of input-h1] {};
		\node[particle] (input-v2) [above =of input-h2] {};
		\node[particle] (input-v3) [above =of input-h3] {};
		
		\path[black,semithick] (input-v1) edge  (input-h1);
		\path[black] (input-v1) edge  (input-h2);
		\path[black] (input-v1) edge  (input-h3);
		\path[black] (input-v2) edge  (input-h1);
		\path[black,semithick] (input-v2) edge  (input-h2);
		\path[black] (input-v2) edge  (input-h2);
		\path[black] (input-v3) edge  (input-h1);
		\path[black] (input-v3) edge  (input-h2);
		\path[black,semithick] (input-v3) edge  (input-h3);
		
		\draw (-.625, -.75) rectangle (2.625+.75, .75);

		\node[treenode] (output-h1) [below =of input-h1] {};
		\node[treenode] (output-h2) [below =of input-h2] {};
		\node[treenode] (output-h3) [below =of input-h3] {};
		\node[treenode] (output-h4) [below =of input-h4] {};
		
		\path[black] (input-h1) edge  (output-h1);
		\path[black] (input-h2) edge  (output-h1);
		\path[black] (input-h3) edge  (output-h1);
		\path[black] (input-h1) edge  (output-h2);
		\path[black] (input-h2) edge  (output-h2);
		\path[black] (input-h3) edge  (output-h2);
		\path[black] (input-h1) edge  (output-h3);
		\path[black] (input-h2) edge  (output-h3);
		\path[black] (input-h3) edge  (output-h3);
		\path[black] (input-h1) edge  (output-h4);
		\path[black] (input-h2) edge  (output-h4);
		\path[black] (input-h3) edge  (output-h4);
		
		\draw ($(output-h1)+(-0.625, -0.75)$) rectangle ($(output-h4)+(0.625, .75)$);
		
		\node[particle] (output-v1) [below =of output-h1] {};
		\node[particle] (output-v2) [below =of output-h2] {};
		\node[particle] (output-v3) [below =of output-h3] {};
		\node[particle] (output-v4) [below =of output-h4] {};
		
		\path[black] (output-h1) edge (output-v1);
		\path[black] (output-h2) edge (output-v2);
		\path[black] (output-h3) edge (output-v3);
		\path[black] (output-h4) edge (output-v4);
		
		\node[] (brace-v1) at ($(output-v1)+(0.4, 0)$) {$\big\}$};
		\node[] (brace-v2) at ($(output-v2)+(0.4, 0)$) {$\big\}$};
		\node[] (brace-v3) at ($(output-v3)+(0.4, 0)$) {$\big\}$};
		
		\draw[dashed, rounded corners] ($(brace-v1)+(0.1,0)$) -| ($(brace-v1)+(0.25,.5)$) |- (output-h2);
		\draw[dashed, rounded corners] ($(brace-v2)+(0.1,0)$) -| ($(brace-v2)+(0.25,.5)$) |- (output-h3);
		\draw[dashed, rounded corners] ($(brace-v3)+(0.1,0)$) -| ($(brace-v3)+(0.25,.5)$) |- (output-h4);

		\node[circle, draw=black!60, line width=0.3mm, fill=white, inner sep=-1] (large-h) at (6.75, 0) {
        	\begin{tikzpicture}[shorten >= 1pt, ->, node distance=.7,scale=0.65]
        		\draw[fill=white, opacity=.25] ($(-1.5, -.9)+(0.3,0.3)$) rectangle ($(1.5, .9)+(0.3,0.3)$);
        		\draw[fill=white, opacity=.5] ($(-1.5, -.9)+(0.2,0.2)$) rectangle ($(1.5, .9)+(0.2,0.2)$);
        		\draw[fill=white, opacity=.75] ($(-1.5, -.9)+(0.1,0.1)$) rectangle ($(1.5, .9)+(0.1,0.1)$);
        		\draw[fill=white] (-1.5, -.9) rectangle (1.5, .9);
        		\node[] (value) at (0.95, 0.35) {V};
        		\node[] (key) at (0.0, 0.35) {K};
        		\node[] (query) at (-0.95, 0.35) {Q};
        		
        		\node[] (in-v) [above= of value.center] {};
        		\node[] (in-k) [above= of key.center] {};
        		\node[] (in-q) [above= of query.center] {};
        		
        		\path[black] (in-v) edge ($(value.center)+(0, 0.3)$);
        		\path[black] (in-k) edge ($(key.center)+(0, 0.3)$);
        		\path[black] (in-q) edge ($(query.center)+(0, 0.3)$);
        		
        		\path[black] (0, -0.9) edge (0, -1.4);
        	\end{tikzpicture}
		};

	\end{tikzpicture}
	
	
	}
    \end{minipage}
    \hfill
	\begin{minipage}{.32\linewidth}
	\resizebox{\linewidth}{!}{
    		\begin{tikzpicture}[shorten >= 1pt, ->, node distance=\layersep,scale=0.65]
 		\tikzstyle{neuron} = [circle, draw=black!60, line width=0.3mm, fill=white];
 		\tikzstyle{treenode} = [circle, draw=black!60, line width=0.3mm, fill=white];
 		\tikzstyle{particle} = [diamond, draw=black!60, line width=0.3mm, fill=white,scale=0.7];

		\node[treenode] (input-h1) at (0,0) {};
		\node[treenode] (input-h2) at (1.25,0) {};
		\node[treenode] (input-h3) at (2.5,0) {};
		
		\node[treenode,opacity=0] (input-h0) at (-1.25,0) {};
		\node[treenode,opacity=0] (input-h4) at (3.75,0) {};
		
		\node[particle] (input-v1) [above =1.2 of input-h1] {};
		\node[particle] (input-v2) [above =1.2 of input-h2] {};
		\node[particle] (input-v3) [above =1.2 of input-h3] {};
		
		\node[particle,opacity=0] (input-v0) [above =1.2 of input-h0] {};
		\node[particle,opacity=0] (input-v4) [above =1.2 of input-h4] {};
		
		\path[black] (input-v1) edge  (input-h1);
		\path[black] (input-v2) edge  (input-h2);
		\path[black] (input-v3) edge  (input-h3);
		
		\draw[-, fill=white] ($(input-v2.center)-(0.1, 0.6)$) -- ($(input-v4.center)-(-0.1, 0.6)$) -- ($(input-h3.center)+(0, 0.6)$) -- cycle;
		\draw[-, fill=white] ($(input-v1.center)-(0.1, 0.6)$) -- ($(input-v3.center)-(-0.1, 0.6)$) -- ($(input-h2.center)+(0, 0.6)$) -- cycle;
		\draw[-, fill=white] ($(input-v0.center)-(0.1, 0.6)$) -- ($(input-v2.center)-(-0.1, 0.6)$) -- ($(input-h1.center)+(0, 0.6)$) -- cycle;

		\draw[-,step=1.25,draw=black!50, line width=0.2mm,xshift=-.625cm,yshift=-.8cm] (0,0) grid (3.75,-5);
		
		\node[particle, opacity=0] (top-of-mat) at (-2, -0.8) {};
		\node[particle, opacity=0] (right-of-mat) at (3.75-0.625, 2) {};
		\node[particle, opacity=0] (left-of-mat) at (-0.625, 2) {};
		\node[particle, opacity=0] (output-x) at (6, 2) {};
		\node[particle, opacity=0] (output2-x) at (-3.75, 2) {};
		\node[particle, opacity=0] (output2-h-x) at (-1.5, 2) {};
		
		\path[black] (input-h1) edge  (input-h1 |- top-of-mat.center);
		\path[black] (input-h2) edge  (input-h2 |- top-of-mat.center);
		\path[black] (input-h3) edge  (input-h3 |- top-of-mat.center);

		\node[] (brace-in-v) at ($(input-v3)+(0.75, 0)$) {$\Big\}$};
		\node[] (brace-in-h) at ($(input-h3)+(0.75, 0)$) {$\Big\}$};
		\node[] (conn-in-v) at ($(input-v3)+(2.35, 0)$) {};
		\node[] (conn-in-h) at ($(input-h3)+(2.35, 0)$) {};
		
		\path[-, shorten >= 0pt] ($(brace-in-v.center)+(0.1,0)$) edge (conn-in-v.center);
		\path[-, shorten >= 0pt] ($(brace-in-h.center)+(0.1,0)$) edge (conn-in-h.center);
	
		\node[particle, opacity=0] (conn1) at (0, -.8-.625-0*1.25) {};
		\node[particle, opacity=0] (conn2) at (0, -.8-.625-1*1.25) {};
		\node[particle, opacity=0] (conn3) at (0, -.8-.625-2*1.25) {};
		\node[particle, opacity=0] (conn4) at (0, -.8-.625-3*1.25) {};
	
		\path[-, shorten >= 0pt] (conn-in-v.center) edge  ($(conn-in-v.center |- conn1.center)+(0,0.1)$);
		\path[-, shorten >= 0pt] ($(conn-in-v.center |- conn1.center)-(0,0.1)$) edge  ($(conn-in-v.center |- conn2.center)+(0,0.1)$);
		\path[-, shorten >= 0pt] ($(conn-in-v.center |- conn2.center)-(0,0.1)$) edge  ($(conn-in-v.center |- conn3.center)+(0,0.1)$);
		\path[-, shorten >= 0pt] ($(conn-in-v.center |- conn3.center)-(0,0.1)$) edge  ($(conn-in-v.center |- conn3.center)-(0,0.1+0.5)$);

		\node[treenode] (out-h1) at ($(conn-in-v.center |- conn1.center)+(-.6,0.)$) {};
		\node[treenode] (out-h2) at ($(conn-in-v.center |- conn2.center)+(-.6,0.)$) {};
		\node[treenode] (out-h3) at ($(conn-in-v.center |- conn3.center)+(-.6,0.)$) {};
		\node[treenode] (out-h4) at ($(conn-in-v.center |- conn4.center)+(-.6,0.)$) {};

		\draw[, rounded corners] ($(conn-in-v.center |- conn1.center)-(0,0.1+0.3)+(0,1.25)$) -| (out-h1);
		\draw[, rounded corners] ($(conn-in-v.center |- conn1.center)-(0,0.1+0.3)$) -|  (out-h2);
		\draw[, rounded corners] ($(conn-in-v.center |- conn2.center)-(0,0.1+0.3)$) -|  (out-h3);
		\draw[, rounded corners] ($(conn-in-v.center |- conn3.center)-(0,0.1+0.3)$) -|  (out-h4);

		\path[] (right-of-mat |- out-h1) edge (out-h1);
		\path[] (right-of-mat |- out-h2) edge (out-h2);
		\path[] (right-of-mat |- out-h3) edge (out-h3);
		\path[] (right-of-mat |- out-h4) edge (out-h4);
		
		\node[particle] (out-v1) at (output-x |- out-h1) {};
		\node[particle] (out-v2) at (output-x |- out-h2) {};
		\node[particle] (out-v3) at (output-x |- out-h3) {};
		\node[particle] (out-v4) at (output-x |- out-h4) {};
		
		\path[] (out-h1) edge (out-v1);
		\path[] (out-h2) edge (out-v2);
		\path[] (out-h3) edge (out-v3);
		\path[] (out-h4) edge (out-v4);

		\node[particle, opacity=0] (out2-v-2) at ($(output2-x.center |- conn1.center)+(0, 2*1.25)$) {};
		\node[particle, opacity=0] (out2-v-1) at ($(output2-x.center |- conn1.center)+(0, 1.25)$) {};
		\node[particle, opacity=0] (out2-v0) at ($(output2-x.center |- conn1.center)$) {};
		
		\node[particle, opacity=.9] (out2-v1) at ($(output2-x.center |- conn2.center)$) {};
		\node[particle, opacity=.9] (out2-v2) at ($(output2-x.center |- conn3.center)$) {};
		\node[particle, opacity=.9] (out2-v3) at ($(output2-x.center |- conn4.center)$) {};
		\node[particle, opacity=.9] (out2-v4) at ($(output2-x.center |- conn4.center)-(0, .75)$) {};
		
		\node[treenode] (out2-h1) at (output2-h-x |- out-h1) {};
		\node[treenode] (out2-h2) at (output2-h-x |- out-h2) {};
		\node[treenode] (out2-h3) at (output2-h-x |- out-h3) {};
		\node[treenode] (out2-h4) at (output2-h-x |- out-h4) {};
		
		\node[rotate=-90] (bottom-brace) at ($(out-v4)+(0,-0.65)$) {$\Big\}$};
		\draw[dashed, rounded corners, shorten >= 5pt] ($(bottom-brace.center)-(0,0.15)$) |- (-3, -6.5) -- (out2-v4);
		
		\path[] ($(out2-h1.center)-(0.6, 0)$) edge (out2-h1);
		\path[] (out2-v1) edge (out2-h2);
		\path[] (out2-v2) edge (out2-h3);
		\path[] (out2-v3) edge (out2-h4);
		
		\draw[-, fill=white, dashed] ($(out2-v-2.center)+(0.6, .0)$) -- ($(out2-v0.center)+(0.6, -.3)$) -- ($(out2-h1.center)-(0.6, 0)$) -- cycle;
		\draw[-, fill=white]  ($(out2-v1.center)+(0.6, .0)$) -- ($(out2-v3.center)+(0.6, -.3)$) -- ($(out2-h4.center)-(0.6, 0)$) -- cycle;
		\draw[-, fill=white]  ($(out2-v0.center)+(0.6, .0)$) -- ($(out2-v2.center)+(0.6, -.3)$) -- ($(out2-h3.center)-(0.6, 0)$) -- cycle;
		\draw[-, fill=white] ($(out2-v-1.center)+(0.6, .0)$) -- ($(out2-v1.center)+(0.6, -.3)$) -- ($(out2-h2.center)-(0.6, 0)$) -- cycle;
	    
		\path[] (out2-h1) edge (out2-h1 -| left-of-mat);
		\path[] (out2-h2) edge (out2-h2 -| left-of-mat);
		\path[] (out2-h3) edge (out2-h3 -| left-of-mat);
		\path[] (out2-h4) edge (out2-h4 -| left-of-mat);

	\end{tikzpicture}
	}
    \end{minipage}
	\caption{Schema of recurrent (top left), recursive (bottom left), transformer (middle), and convolutional  sequence-to-sequence (right) neural networks.}
	\label{fig:nn-schemas}
\end{figure*}
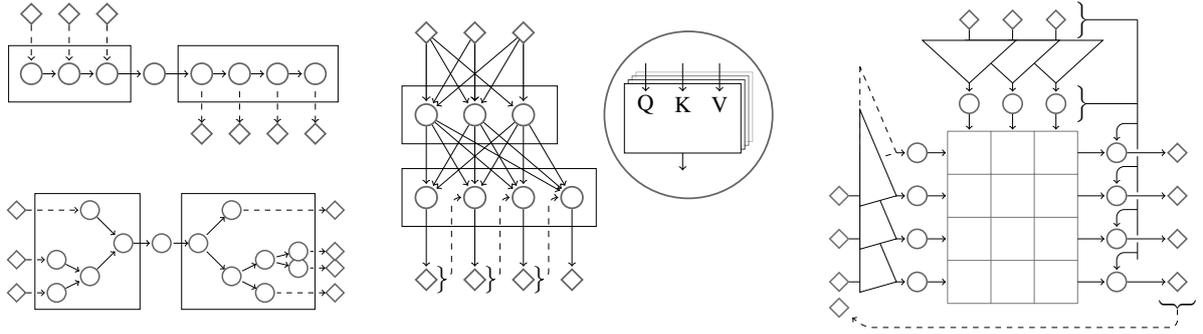

Mathematical formulae are generally similar to natural language \cite{Greiner-PetterS20}.
However, mathematical formulae are often much longer than natural language sentences.
As an example of sentence lengths, $98\%$ of the sentences in the Stanford Natural Language Inference entailment task contain less than 25 words \cite{Bowman2016}.
In contrast, the average number of Mathematica tokens in the Mathematical Functions Site data set is $173$, only $2.25\%$ of the formulae contain less than 25 tokens, and $2.1\%$ of the formulae are longer than $1\,024$ tokens.
At the same time, mathematical languages commonly require only small vocabularies of around $1\,000$ tokens (relative to natural languages.)

By applying convolutional sequence-to-sequence networks, this work achieves an exact match accuracy of $95.1\%$ for a translation from \LaTeX{} to Mathematica as well as an accuracy of $90.7\%$ for a translation from \LaTeX{} to \textit{semantic} \LaTeX{}.
In contrast, the import function of the Mathematica software achieves an exact match accuracy of $2.7\%$.
On all measured metrics, our model outperforms export / import round trips using Mathematica.

\section{Related Work}
\label{sec.related_work}

\subsection{Neural Machine Translation}
\label{sec:nmt}

The most common neural machine translation models are sequence-to-sequence recurrent neural networks \cite{Sutskever2014Seq2Seq}, tree-structured recursive neural networks \cite{Goller2002}, transformer sequence-to-sequence networks \cite{Vaswani2017}, and convolutional sequence-to-sequence networks \cite{Gehring2017}.
In the following, we sketch the core principle of these network types, which are displayed in Figure~\ref{fig:nn-schemas}.

Recurrent sequence-to-sequence neural networks (Figure~\ref{fig:nn-schemas}, top left) are networks that process the tokens one after each other in a linear fashion.
Note that the longest shortest path in this architecture is the sum of the length of the input and the length of the output.
An attention mechanism can reduce the loss of information in the network (not shown in the schema). 

Recursive tree-to-tree neural networks (Figure~\ref{fig:nn-schemas}, bottom left) are networks that process the input in a tree-like fashion.
Here, the longest shortest path is the sum of the depths of input and output, i.e., logarithmic in the number of tokens.

Transformer sequence-to-sequence neural networks (Figure~\ref{fig:nn-schemas}, middle) allow a dictionary-like lookup of hidden states produced from the input sequence.
This is possible through an elaborate multi-headed attention mechanism.

Convolutional sequence-to-sequence neural networks (Figure~\ref{fig:nn-schemas}, right) process the input using a convolutional neural network and use an attention-mechanism to attribute which input is most relevant for predicting the next token given previously predicted tokens.

In natural language translation, transformer networks perform best, convolutional second best, and recurrent third best \cite{Gehring2017,Vaswani2017,Ott2018}.
Recursive neural networks are commonly not applicable to natural language translation.

\subsection{Rule-Based Formula Translation}\label{sec:ruleform}

\LaTeX{}ML is a \LaTeX{} to XML Converter \cite{Ginev2014}. 
It can translate from \textit{semantic} \LaTeX{} to \LaTeX{}.
As semantic information is lost during this process, a rule-based back-translation is not possible.

Mathematica can export expressions into \LaTeX{} and also import from \LaTeX{}.
However, the import from \LaTeX{} uses strict and non-exhaustive rules that oftentimes do not translate into the original Mathematica expressions, e.g., we found that only $3.1\%$ of expressions exported from Mathematica to \LaTeX{} and (without throwing an error) imported back into Mathematica are exact matches.
This is because, when translating into \LaTeX{}, the semantic information is lost.
Moreover, we found that $11.5\%$ of the formulae exported from Mathematica throw an error when reimporting them.

For the translation between CLs, from \textit{semantic} \LaTeX{} to CASs and back, there exists a rule-based translator~\cite{Cohl17,GreinerPetter2019-05-20Seman-46001}.
The \textit{semantic} \LaTeX{} to Maple translator achieved an accuracy of $53.59\%$ on correctly translating $4\,165$ test equations from the DLMF~\cite{GreinerPetter2019-05-20Seman-46001}.
The accuracy of the \textit{semantic} \LaTeX{} to CAS translator is relatively low due to the high complexity of the tested equations and because many of the functions which are represented by a DLMF/DRMF \LaTeX{} macro are not defined or defined differently in Maple~\cite{GreinerPetter2019-05-20Seman-46001}.

\subsection{Deep Learning for Mathematics}
\label{sec:nmt.math}

\citet{lample2019deep} used deep learning to solve symbolic mathematics problems.
They used a sequence-to-sequence transformer model to translate representations of mathematical expressions into representations of solutions to problems such as differentiation or integration.
In their results, they outperform CASs such as Mathematica.

\citet{Wang2018Mizar} used a recurrent neural network-based sequence-to-sequence model to translate from \LaTeX{} (text including formulae) to the Mizar language, a formal language for writing mathematical definitions and proofs.
Their system generates correct Mizar statements for $65.7\%$ of their synthetic data set.

Other previous works~\cite{Deng2016,Wang2019} concentrated on the ``image2latex'' task, which was originally proposed by OpenAI.
This task's concept is the conversion of mathematical formulae in images into \LaTeX{}, i.e., optical character recognition of mathematical formulae.
\citet{Deng2016} provide im2latex-100k, a data set consisting of about $100\,000$ formulae from papers of arXiv, including their renderings.
They achieved an accuracy of $75\%$ on synthetically rendered formulae.
Compared to the data sets used in this work, the formulae in im2latex-100k are much shorter.

This was followed by other relevant lines of work by \citet{DBLP:conf/acl/WuZWH20,DBLP:conf/acl/ZhangWLBWSL20,li-etal-2022-seeking,ferreira-etal-2022-integer,patel2021nlp}.

\section{Training Data Sets \& Preprocessing}
\label{apx.data}

\paragraph{Mathematical Functions Site Data Set.}
\label{apx.data.mathematica}
The ``Mathematical Functions Site''\footnote{\url{http://functions.wolfram.com/}} by Wolfram Research is a repository of $307\,672$ mathematical formulae available in Mathematica InputForm format.
By web-crawling, we retrieved all formulae in Mathematica InputForm and (using Mathematica v12.0) exported the formulae from Mathematica into \LaTeX{}.

\paragraph{\textit{Semantic} \LaTeX{} Data Set.}
\label{apx.data.semanticlatex}
The \textit{semantic} \LaTeX{} data set consist of $11\,639$ pairs of formulae in the \LaTeX{} and \textit{semantic} \LaTeX{} formats generated by translating from \textit{semantic} \LaTeX{} to \LaTeX{} using \LaTeX{}ML.
\citet{Cohl15} provided us this unreleased data set.

\paragraph{Preprocessing}
\label{apx.preprocessing}
We preprocessed the data sets by tokenizing them with custom rule-based tokenizers for \LaTeX{} and Mathematica.
Note that as \textit{semantic} \LaTeX{} follows the rules of \LaTeX{}, we can use the same for both cases.
Details on the tokenizers are presented in the supplementary material.
For recursive neural networks, we parsed the data into respective binary trees in postfix notation.

We randomly split the Mathematical Functions Site data set into disjoint sets of $97\%$ training, $0.5\%$ validation, and $2.5\%$ test data and split the \textit{semantic} \LaTeX{} data set into $90\%$ training, $5\%$ validation, and $5\%$ test data since this data set is smaller.
Data set summary statistics can be found in Table~\ref{dataset-stats}.

\begin{table}[]
	\centering
	\caption{
		Data set summary statistics. Format for number of characters per formula/format: Mean$\pm$Std.~(Median).
	}
	\resizebox{\linewidth}{!}{
		\begin{tabular}{lrcccccccc}
			\toprule
			Data Set                     & \kern-1emFormulae & \kern-.5emInput (\LaTeX{}) & \kern-1.5emOutput (Mat. / \textit{sem.}~L.) \\
			\midrule
			Mathematica                  & $307\,672$ & $345.5\pm534.4~(195)$ & $320.7\pm585.7~(168)$  \\
			\textit{semantic}~\LaTeX{}   & $11\,639$ & $163.8\pm246.2~(116)$ & $145.6\pm230.1~(103)$   \\
			\bottomrule
		\end{tabular}
		}
	\label{dataset-stats}
\end{table}

\section{Methods}
\label{sec.method}

We briefly discuss recurrent, recursive, and transformer architectures and then discuss convolutional sequence-to-sequence networks in detail because they showed, by far, the best results.

\paragraph{Recurrent Neural Networks} showed the worst performance.
Our experiments used Long-Short-Term-Memory (LSTM) recurrent networks but did not achieve any exact matches on long equations of the \textit{semantic} \LaTeX{} data set.
This is not surprising as recurrent neural networks generally have poor performance regarding long-term relationships spanning over hundreds of tokens \cite{trinh2018learning}.
For our data sets, the longest shortest path in the neural network easily exceeds $2\,000$ blocks.
Note that the exact match accuracy on such long equations produces successful responses only for a very well-performing model; getting most symbols correct does not constitute an exact match. 
For a definition of exact matches, see Section~\ref{sec.metrics}.

\paragraph{Recursive Neural Networks} showed slightly better performance of up to $4.4\%$ exact match accuracy when translating from \LaTeX{} into \textit{semantic} \LaTeX{}.
This can be attributed to the fact that the longest path inside a recursive neural network is significantly shorter than in a recurrent neural network (as the longest shortest path in a tree can be much shorter than the longest shortest path in a sequence.)
Further, an additional traversal into postfix notation allows for an omission of most braces/parentheses, which (on the \textit{semantic} \LaTeX{} data set) reduced the required amount of tokens per formula by about $20-40\%$.
Similar to the recurrent networks, we also used LSTMs for the recursive networks.
Note that training recursive neural networks is hard because they cannot easily be batched if the topology of the trees differs from sample to sample, which it does for equations.

\paragraph{Transformer Neural Networks} significantly outperform previous architectures.
In our best experiments, we achieved performances of up to $50\%$ exact matches on the Mathematical Functions Site data set.
This leap in performance can be attributed to the elaborate multi-headed attention mechanism underlying the transformer model.
Because we experimented simultaneously with the convolutional sequence-to-sequence architecture and the transformer architecture, and the performance of convolutional networks was by a large margin better ($>90\%$) than the best performance on transformer neural networks, we decided to set the focus of this work on convolutional networks only.
We note that in natural language translation, transformer models typically outperform convolutional neural networks~\cite{Gehring2017,Vaswani2017,Ott2018}.

\subsection{Convolutional Seq-to-Seq Networks}
\label{sec:convolutional_seq}

In contrast to recurrent and recursive neural networks, convolutional sequence-to-sequence networks do not need to compress the relevant information.
Due to the attention matrix architecture, the convolutional model can easily replicate the identity, a task that recurrent and recursive neural networks struggle with.
In fact, an above $99\%$ accuracy can be achieved on learning the identity within the first epoch of training.
Given that the syntax of two languages follows the same paradigm, the translation is often not far from the identity, e.g., it is possible that only some of the tokens have to be modified while many remain the same.
This separates mathematical notations from natural languages.

In the following, we discuss hyperparameters and additional design choices for convolutional networks.
Note that the models for each language pair are independent. 
In Supplementary Material~\ref{ablation}, we provide respective ablation studies.

\paragraph{Learning Rate, Gradient Clipping, Dropout, and Loss.}
Following the default for this model, we use a learning rate of $0.25$, applied gradient clipping on gradients greater than $0.1$, and used a dropout rate of $0.2$.
As a loss, we use label-smoothed cross-entropy.

\paragraph{State/Embedding Size(s).} 
We found that a state size of 512 performs best.
In this architecture, it is possible to use multiple state sizes by additional fully connected layers between convolutional layers of varying state size.
In contrast to the convolutional layers, fully connected layers are not residual and thus increase the length of the shortest path in the network.
We found that networks with a single state size performed best.
Note that while in natural language translation, with vocabularies of $40\,000-200\,000$ tokens, a state size of 512 is also commonly used \cite{Gehring2017}, while our examined mathematical languages contain only $500-1\,000$ tokens.
That a state size of 256 performed significantly worse ($88.3\%$ for 256 and $94.9\%$ for 512) indicates a high entropy/information content of the equations.

\paragraph{Number of Layers.} We found that 11 layers perform best.

\paragraph{Batch Size.}
We found that $48\,000$ tokens per batch perform best.
This is equivalent to a batch size of about 400 formulae.

\paragraph{Kernel Size.}
We use a kernel size of 3.
We found that a kernel size of 5 performs by $0.1\%$ better than a kernel size of 3, but as the larger kernel size also requires much more parameters and is more expensive to compute, we decided to go with 3.

\paragraph{Substitution of Numbers.}
\label{sec.substitutenumbers}
Since the Mathematical Functions Site data set contains more than $10^4$ multi-digit numbers, while it contains less than $10^3$ non-numerical tags, these numbers cannot be interpreted as conventional tags.
Thus, numbers are either split into single digits or replaced by variable tags.
Splitting numbers into single digits causes significantly longer token streams, which degrades performance.
Substituting all multi-digit numbers with tags like \verb|<number_01>| improved the exact match accuracy of the validation data set from $92.7\%$ to $95.0\%$.
We use a total of 32 of such placeholder tags as more than $99\%$ of the formulae have less or equal to $32$ multi-digit numbers.
We randomly select the tags that we substitute the numbers with.
Since multi-digit numbers basically always perfectly correspond in the different mathematical languages, we directly replace the tag with their corresponding numbers after the translation.

\paragraph{LightConv.}
As an alternative to the model proposed by \citet{Gehring2017}, we also used the LightConv model as presented by \citet{Wu2019PayLessAttention}.
As expected, this model did not yield good results on mathematical formula translation as it does not use the strong self-attention that the model by \citet{Gehring2017} has.
Note that LightConv outperforms the convolutional sequence-to-sequence model by \citet{Gehring2017} on natural language~\cite{Wu2019PayLessAttention}.

\section{Evaluation of the Convolutional Network}
\label{sec.results}

\subsection{Evaluation Metrics}
\label{sec.metrics}

\paragraph{Exact Match (EM) Accuracy} The EM accuracy is the non-weighted share of exact matches.
An exact match is defined as a translation of a formula where every token equals the ground truth.
This makes the EM accuracy an extremely strict metric as well as a universal and definite statement about a lower bound of the quality of the translation.
For example, the exact match might fail since $E=mc^2$ can be written as both \textbf{\texttt{E=mc\^{}2}} and \textbf{\texttt{E=mc\^{}\{2\}}}, which is, although content-wise equal, not an exact match.
However, in our experiments, such errors do not occur regularly since, for the generation of the synthetic training data, the translation was performed using the rule-based translators Mathematica and \LaTeX{}ML.
Only $0.4\%$ of the erroneous translations to \textit{semantic} \LaTeX{} were caused by braces (\textbf{\texttt{\{}}, \textbf{\texttt{\}}}).
In none of these cases the braces were balanced, i.e., each of these formulae was semantically incorrect.
For the translation to Mathematica, only $0.02\%$ of the formulae did not achieve an exact match due to brackets (\textbf{\texttt{[}}, \textbf{\texttt{]}}).

\paragraph{Levenshtein Distance (LD)} The LD, which is also referred to as ``edit distance'', is the minimum number of edits required to change one token stream into another \cite{Levenshtein1966}.
This metric reflects the error in a more differentiated way.
We denote the share of translations that have a Levenshtein distance of up to $5$ by $\mathrm{LD}_{\leq 5}$ and denote the average Levenshtein Distance by $\mathrm{LD}$.

\paragraph{Bilingual Evaluation Understudy (BLEU)} The BLEU score is a quality measure that compares the machine's output to a translation by a professional human translator \cite{Papineni2002-bleu}.
It compares the $n$-grams (specifically $n\in\{1,2,3,4\}$) between the prediction and the ground truth.
Since the translations in the data sets are ground truth values instead of human translations, for the back-translation of formulae, this metric reflects the closeness to the ground truth.
BLEU scores range from 0 to 100, with a higher value indicating a better result.
For natural language on the WMT data set, state-of-the-art BLEU scores are $35.0$ for a translation from English to German and $45.6$ for a translation from English to French \cite{Edunov2018}.
That the BLEU scores for formula translations are significantly higher than the scores for natural language can be attributed to the larger vocabularies in natural language and a considerably higher variability between correct translations.
In contrast, in most cases of formula translation, the translation is not ambiguous.
We report the BLEU scores to demonstrate how BLEU scores behave on strictly defined languages like mathematical formulae.

\paragraph{Perplexity} The perplexity is a measurement of how certain a probability distribution is to predict a sample.
Specifically, the perplexity of a discrete probability distribution $p$ is generally defined as 
\begin{equation}
	\texttt{ppl} (p) = 2^{H(p)} = 2^{-\sum_{x} p(x) \log_2 p(x)}
\end{equation}
where $H$ denotes the entropy, and $x$ is drawn from the set of all possible translations \cite{Cover2006EIT}.
In natural language processing, a lower perplexity indicates a better model.
As we will discuss later, this does not hold for mathematical language.

\subsubsection{Discussion on the Perplexity of Mathematical Language Translations}
\label{sec.discuss_perplexity}

In natural language translation, the perplexity is a common measure for selecting the epoch at which the performance on the validation set is best.
That is because its formulation is very similar to the employed cross-entropy loss.
This procedure avoids overfitting and helps to select the best-performing epoch without having to compute the actual translations.
Computing the translations would be computationally much more expensive because it requires a beam search algorithm, and the quality of a resulting translation cannot be measured by a simple metric such as EM.

However, for formula translation, the perplexity does not reflect the accuracy of the model.
While the validation accuracy rises over the course of the training, the rising perplexity falsely indicates that the model's performance decays during training.
We presume that this is because the perplexity reflects how sure the model is about the prediction instead of whether the prediction with the highest probability is correct.
Since many subexpressions of mathematical formulae (e.g., \texttt{n + 1}) are invariant to translations between many mathematical languages, the translations are closer to the identity than translations between natural languages.
Therefore, a representation very close to the identity is learned first.
Consecutively, this translation is transformed into the actual translation.
Empirically, the validation perplexity usually reaches its minimum during the first epoch.
Afterward, when the translation improves, the uncertainty (perplexity) of the model also increases.
Thus, we do not use the perplexity for early stopping but instead compute the EM accuracy on the validation set.

\subsection{Evaluation Techniques}

\paragraph{Back-Translation.}
As, for the training data sets, only the content language (i.e., Mathematica / \textit{semantic} \LaTeX{}, respectively) was available, we programmatically generated the input forms (presentation language) using Mathematica's conversion and the \LaTeX{} macro definitions of \textit{semantic} \LaTeX{}, respectively. 
This process corresponds to the internal process for displaying Mathematica / \textit{semantic} \LaTeX{} equations in \LaTeX{} form.
Thus, the task is to back-translate from (ambiguous) \LaTeX{} to the (unambiguous) Mathematica / \textit{semantic} \LaTeX{} forms.

\paragraph{Additional Experiments.}
In addition to this, we also perform round trip experiments from \LaTeX{} into Mathematica and back again on the im2latex-100k data set.
Here, we use our model as well as the Mathematica software to translate from \LaTeX{} into Mathematica.
In both cases, we use Mathematica to translate back into \LaTeX{}.
The im2latex-100k data set contains equations as well as anything else that was typeset in math environments \LaTeX{}.
$66.8\%$ of the equations in the im2latex-100k data set contain tokens that are not in the vocabulary.
We note that an exact match is only possible if a \LaTeX{} expression coincides with what would be exported from Mathematica.
Thus, we did not expect large accuracy values for this data set.

\begin{table}[]
\newcommand{\res}[3]{$#1\%$ | $#2$}
	\centering
	\caption{
		Main results for the back-translation. 
	}
    \vspace{-.5em}
	\label{tab:em-bleu}
	\resizebox{\linewidth}{!}{
	\begin{tabular}{ l c c}
		\toprule
		Metric & \LaTeX{} $\to$ Mathematica & \LaTeX{} $\to$ \textit{semantic} \LaTeX{} \\
		\midrule
		EM 			& 95.1\% & 90.7\% \\
		BLEU 			& {99.68} & {96.79} \\
		\bottomrule
	\end{tabular}
	}
\end{table}

\newcommand{\canInterpret}[0]{{Valid Mathe\-mati\-ca}\hspace*{-5.25em}}
\newcommand{\canInterpretM}[0]{\parbox[c]{7em}{Can import}\hspace*{-6.25em}}
\newcommand{\exactMatch}[0]{EM }
\newcommand{\bleuScore}[0]{BLEU }
\newcommand{\LDfive}[0]{$\mathrm{LD}_{\leq5}$}
\newcommand{\LDavg}[0]{$\mathrm{LD}$}

\begin{table}[]
	\centering
	\caption{
		Comparison between Mathematica and our model on back-translating the formulae of the Mathematical Functions Site data set.
		Import denotes the fraction of formulae that can be imported by Mathematica, i.e., whether Mathematica can import the \LaTeX{} format or whether our model produces valid Mathematica syntax, respectively.
	}
    \vspace{-.5em}
	\resizebox{\linewidth}{!}{
		\begin{tabular}{ l  c  c  c  c }
			\toprule
			Method          & \exactMatch & Import & \LDfive & \LDavg \\
			\midrule
			Mathematica     & $2.7\%$ & $88.5\%$ & $16.4\%$ & $88.7$ \\
			Conv.~Seq2Seq   & $\mathbf{95.1\%}$ & $\mathbf{98.3\%}$ & $\mathbf{96.7\%}$ & $\mathbf{0.615 }$ \\
			\bottomrule
		\end{tabular}
		}
	\label{mathematica-results}
\end{table}

\subsection{Evaluation Results}

\paragraph{Back-Translation.}
For the back-translation from \LaTeX{} to Mathematica, we achieved an EM accuracy of $95.1\%$ and a BLEU score of $99.68$.
That is, $95.1\%$ of the expressions from Mathematica, translated by Mathematica into \LaTeX{} can be translated back into Mathematica by our model without changes.
For the translation from \LaTeX{} to \textit{semantic} \LaTeX{}, we achieved an EM accuracy of $90.7\%$ and a BLEU score of $96.79$.
The translation from \LaTeX{} to \textit{semantic} \LaTeX{} performs not as well as the translation to Mathematica, i.a., because the \textit{semantic} \LaTeX{} data set is substantially smaller than the Mathematical Functions Site data set.
The low \LaTeX{} to \textit{semantic} \LaTeX{} BLEU score of only $96.79$ is because the translations into \textit{semantic} \LaTeX{} are on average $2\%$ shorter than the ground truth references.
Note that $96.0\%$ of the translations to \textit{semantic} \LaTeX{} had an LD of up to 3.
The results are displayed in Table~\ref{tab:em-bleu}.
\\
For comparing our model to the \LaTeX{} import function of Mathematica, we show the results in Table~\ref{mathematica-results}.
The low performance of Mathematica's \LaTeX{} importer can be attributed to the fact that Symbols with a defined content/meaning, e.g., \texttt{DiracDelta} are exported to \LaTeX{} as \texttt{\textbackslash{}delta}, i.e., just as the character they are presented by.
Since \texttt{\textbackslash{}delta} is ambiguous, Mathematica interprets it as \texttt{\textbackslash{}[Delta]}.
With neural machine translation, on the other hand, the meaning is inferred from the context and, thus, it is properly interpreted as \texttt{DiracDelta}.

\paragraph{Additional Experiments.}
As for the round trip experiments, Mathematica was able to import $15.3\%$ of the expressions in the im2latex-100k data set, while our model was able to generate valid Mathematica syntax for $16.3\%$ of those expressions.
For the im2latex-100k data set, the round trip experiment is ill-posed since the export to \LaTeX{} will only achieve an exact match if the original \LaTeX{} equation is written in the style in which Mathematica exports.
However, as the same Mathematica export function is used for testing for exact matches, neither our model nor the Mathematica software has an advantage on this problem, which allows for a direct comparison.
Mathematica achieved an exact match round trip in $0.153\%$ and our model in $0.698\%$ of the equations.
The average LD for Mathematica is $18.3$, whereas it is $12.9$ for our model.
We also note that while im2latex-100k primarily contains standard equations, our model is specifically trained to interpret equations with special functions.
The results are presented in Table~\ref{round-trip-im2latex-100k}.

\subsection{Qualitative Analysis}

We present a qualitative analysis of the back-translations from \LaTeX{} to Mathematica with the help of randomly selected positive and negative examples.
The referenced translations / equations are in the supplementary material.
All mentioned parts of equations will be marked in bold in the supplementary material.
We want to give a small qualitative analysis of the translation from \LaTeX{} to Mathematica and show in which cases the translation can fail, and give an intuition about why issues arise in these cases.
In the supplementary material, further qualitative analysis is provided.

In Equation B.1, $\sigma_k(n)$ is correctly interpreted by our model as a \texttt{DivisorSigma}.
Mathematica interprets it as the symbol $\sigma$ with the subscript $k$, i.e., the respective semantic information is lost.
At the end of this formula, the symbol $\land$ (\texttt{\textbackslash{}land}) is properly interpreted by our model as \texttt{\&\&}.
In contrast, Mathematica interpreted it as \texttt{\textbackslash{}[Wedge]}, which corresponds to the same presentation but without the underlying definition that is attached to \texttt{\&\&}.
In this equation, our approach omitted one closing bracket at a place where two consecutive closing brackets should have been placed.

In Equation B.2, the symbol $\wp$ (\texttt{\textbackslash{}wp}) is properly interpreted by the model and Mathematica as the Weierstrass' elliptic function $\wp$ (\texttt{WeierstrassP}).
That is because the symbol $\wp$ is unique to the Weierstrass $\wp$ function.
The inverse of this function, $\wp^{-1}$ is also properly interpreted by both systems as the \texttt{InverseWeierstrassP}.
Our model correctly interprets the sigmas in the same equation as the \texttt{WeierstrassSigma}.
As $\sigma$ does not have a unique meaning, Mathematica just interprets it as a bare sigma \texttt{\textbackslash{}[Sigma]}.
The difference between our translation and the ground truth is that our translation omitted a redundant pair of parentheses.

Equation B.3 displays an example of the token \texttt{<number\_XX>}, which operates as a replacement for multi-digit numbers.
In this example, our model interprets $Q_4^9(z)$ as \texttt{GammaRegularized[4, 9, z]} instead of the ground truth \texttt{LegendreQ[4, 9, 3, z]}.
This case is especially hard since the argument ``\texttt{3}'' is not displayed in the \LaTeX{} equation and \texttt{LegendreQ} has commonly only two to three arguments.

\begin{table}[]
	\centering
	\caption{
		Round trip experiment with the im2latex-100k~\cite{Deng2016} \LaTeX{} expressions.
		Import denotes the fraction of formulae that can be imported by Mathematica, i.e., whether Mathematica can import the \LaTeX{} format or whether our model produces valid Mathematica syntax, respectively.
		In this experiment, exact matches can only occur coincidental, i.e., a perfect translation by the model does not necessarily produce an exact match. 
	}
    \vspace{-.5em}
	\resizebox{\linewidth}{!}{
		\begin{tabular}{ l  c  c  c  c }
			\toprule
			Method          & \exactMatch & Import & \LDfive & \LDavg \\
			\midrule
			Mathematica     & $0.153\%$ & $15.3\%$ & $2.30\%$ & $18.3$ \\
			Conv.~Seq2Seq   & $\mathbf{0.698\%}$ & $\mathbf{16.3\%}$ & $\mathbf{2.56\%}$ & $\mathbf{12.9}$ \\
			\bottomrule
		\end{tabular}
		}
	\label{round-trip-im2latex-100k}
\end{table}

Equation B.7 is correctly interpreted by our model including the expression \texttt{\textbackslash{}int \textbackslash{}sin (az) ... dz}.
Note that Mathematica fails at interpreting \texttt{\textbackslash{}int 2z dz}\footnote{The command \texttt{ToExpression["\textbackslash{}\textbackslash{}int 2z dz", TeXForm, Defer]} fails for Mathematica v.~12.0}.\\

To test whether our model can perform translations on a data set that was generated by a different engine, we perform a manual evaluation on translations from \LaTeX{} to Mathematica for the DLMF data set (generated by \LaTeX{}ML).
To test our model, which was trained on \LaTeX{} expressions produced by Mathematica, on \LaTeX{} expressions produced by \LaTeX{}ML, we used a data set of $100$ randomly selected expressions from the DLMF, which is written in \textit{semantic} \LaTeX{}. 
A caveat of this is that \LaTeX{}ML produces a specific \LaTeX{} flavor in which some mathematical expressions are denoted in an unconventional fashion\footnote{For example, \LaTeX{}ML denotes the binomial as \texttt{\textbackslash{}genfrac\{(\}\{)\}\{0pt\}\{0\}\{n\}\{k\}} instead of \texttt{\textbackslash{}binom\{n\}\{k\}}}. %
As $71$ of those $100$ expressions contain tokens that are not in the Mathematica-export vocabulary, these cannot be interpreted by the model.
Further, as \LaTeX{} is very flexible, a large variety of \LaTeX{} expressions can produce a visually equivalent result; even among a restricted vocabulary, there are many equivalent \LaTeX{} expressions.
This causes a significant distributional domain shift between \LaTeX{} expressions generated by different systems.
Our model generates valid and semantically correct Mathematica representations for 5 equations. 
Specifically, in equations (4.4.17), (8.4.13), and (8.6.7), the model was able to correctly anticipate the incomplete Gamma function and Euler's number $e$. 

This translation from DLMF to Mathematica is difficult for several reasons as explained by \citet{GreinerPetter2019-05-20Seman-46001}.
In their work, they translate the same 100 equations, however, from \textit{semantic} \LaTeX{} into Mathematica, using their rule-based translator, which was designed for this specific task~\cite{GreinerPetter2019-05-20Seman-46001}.
On this different task, they achieved an accuracy of only $56\%$, which clearly shows how difficult a translation between two systems is even when the semantic information is explicitly provided by \textit{semantic} \LaTeX{} expressions.

In comparison, when the vocabulary of \LaTeX{}ML and Mathematica intersects, our model achieves a $17\%$ accuracy while only inferring the implicit semantic information (i.e., the semantic information that can be derived from the structure of and context within a \LaTeX{} expression).

\section{Limitations}

In this work, we evaluated neural networks on the task of back-translating mathematical formulae from the PL \LaTeX{} to semantic CLs.
For this purpose, we explored various types of neural networks and found that convolutional neural networks perform best.
Moreover, we observed that the perplexity of the translation of mathematical formulae behaves differently from the perplexity of the translation between natural languages.

Our evaluation shows that our model outperforms the Mathematica software on the task of interpreting \LaTeX{} produced by Mathematica while inferring the semantic information from the context within the formula.

A general limitation of neural networks is that trained models inherit biases from training data.
For a successful formula translation, this means that the set of symbols, as well as the style in which the formulae are written, has to be present in the training data.
Mathematica exports into a very common flavor / convention of \LaTeX{}, while \textit{semantic} \LaTeX{}, translated by \LaTeX{}ML, yields many unconventional \LaTeX{} expressions.
In both cases, however, the flavor / conventions of \LaTeX{} are constant and do not allow variation as it is produced by a rule-based translator.
Because of the limited vocabularies as well as limited set of \LaTeX{} conventions in the data sets, the translation of mathematical \LaTeX{} expressions of different flavors is not possible.
In addition, we can see that a shift to a more difficult domain, such as special functions in the DLMF, produces a drop in performance but still generates very promising results.
In future work, the translator could be improved by augmenting the data set such that it uses more and different ways to express the same content in the source language.
As an example, a random choice between multiple ways to express a Mathematica expression in \LaTeX{} could be added.
For \textit{semantic} \LaTeX{}, the performance on real-world data could be improved by using multiple macro definitions for each macro.
Ideal would be a data set of hand-written equivalents between the PLs and CLs.
An addition could be multilingual translation \cite{Johnson2017,Blackwood2018}. 
This could allow learning translations and tokens that are not present in the training data for the respective language pair.
Further, mathematical language-independent concepts could support a shared internal representation.

Another limitation is that data sets of mathematical formulae are not publicly available due to copyright and licensing.
We will attempt to mitigate this issue by providing the data sets to interested researchers.

Note that this work does not use information from the context around a formula.
Integrating such context information would aid the translation as it can solve ambiguities.
For example, for interpreting the expression $(x)_n$, information about the specific field of mathematics is essential.
Further, context information can include custom mathematical definitions.
In real-world applications, building on such additional information could be important for reliable translations.

\section{Conclusion}

In this work, we have shown that neural networks, specifically convolutional sequence-to-sequence networks, can handle even long mathematical formulae with high precision.
Given an appropriate data set, we believe that it is possible to train a reliable formula translation system for real-world applications.

We hope to inspire the research community to apply convolutional neural networks rather than transformer networks to tasks that operate on mathematical representations~\cite{Deng2016,DBLP:conf/acl/MatsuzakiIIAA17,lample2019deep,Wang2018Mizar,DBLP:conf/acl/WuZWH20,DBLP:conf/acl/ZhangWLBWSL20,patel2021nlp,li-etal-2022-seeking,ferreira-etal-2022-integer}.
We think that convolutional networks could also improve program-to-program translation as source code has strong similarities to digital mathematical notations---after all, \LaTeX{} and Mathematica are programming languages.

\subsubsection*{Acknowledgments}

This work was supported by the German Academic Exchange Service (DAAD) - 57515245, the Lower Saxony Ministry of Science and Culture, and the VW Foundation.

{
  \bibliography{manualft}

\def\cprime{$'$}
\begin{thebibliography}{36}
\expandafter\ifx\csname natexlab\endcsname\relax\def\natexlab#1{#1}\fi

\bibitem[{Blackwood et~al.(2018)Blackwood, Ballesteros, and
  Ward}]{Blackwood2018}
Graeme Blackwood, Miguel Ballesteros, and Todd Ward. 2018.
\newblock Multilingual neural machine translation with task-specific attention.
\newblock In \emph{Proceedings of the 27th International Conference on
  Computational Linguistics}.

\bibitem[{Bowman et~al.(2016)Bowman, Gauthier, Rastogi, Gupta, Manning, and
  Potts}]{Bowman2016}
Samuel~R. Bowman, Jon Gauthier, Abhinav Rastogi, Raghav Gupta, Christopher~D.
  Manning, and Christopher Potts. 2016.
\newblock A fast unified model for parsing and sentence understanding.
\newblock In \emph{Proceedings of the 54th Annual Meeting of the Association
  for Computational Linguistics (Volume 1: Long Papers)}.

\bibitem[{Cohl et~al.(2015)Cohl, Schubotz, McClain, Saunders, Zou, Mohammed,
  and Danoff}]{Cohl15}
Howard~S Cohl, Moritz Schubotz, Marjorie~A McClain, Bonita~V Saunders, Cherry~Y
  Zou, Azeem~S Mohammed, and Alex~A Danoff. 2015.
\newblock {Growing the Digital Repository of Mathematical Formulae with Generic
  \LaTeX{} Sources}.
\newblock In \emph{Intelligent Computer Mathematics (CICM)}.

\bibitem[{Cohl et~al.(2017)Cohl, Schubotz, Youssef, Greiner-Petter, Gerhard,
  Saunders, McClain, Bang, and Chen}]{Cohl17}
Howard~S. Cohl, Moritz Schubotz, Abdou Youssef, Andr{\'e} Greiner-Petter,
  J{\"u}rgen Gerhard, Bonita~V. Saunders, Marjorie~A. McClain, Joon Bang, and
  Kevin Chen. 2017.
\newblock Semantic preserving bijective mappings of mathematical formulae
  between document preparation systems and computer algebra systems.
\newblock In \emph{Intelligent Computer Mathematics (CICM)}.

\bibitem[{Cover and Thomas(2006)}]{Cover2006EIT}
Thomas~M Cover and Joy~A Thomas. 2006.
\newblock \emph{{Elements of Information Theory (Wiley Series in
  Telecommunications and Signal Processing)}}.
\newblock Wiley-Interscience.

\bibitem[{Dauphin et~al.(2017)Dauphin, Fan, Auli, and Grangier}]{Dauphin2016}
Yann~N. Dauphin, Angela Fan, Michael Auli, and David Grangier. 2017.
\newblock Language modeling with gated convolutional networks.
\newblock In \emph{International Conference on Machine Learning (ICML)}.

\bibitem[{Deng et~al.(2017)Deng, Kanervisto, Ling, and Rush}]{Deng2016}
Yuntian Deng, Anssi Kanervisto, Jeffrey Ling, and Alexander~M. Rush. 2017.
\newblock Image-to-markup generation with coarse-to-fine attention.
\newblock In \emph{International Conference on Machine Learning (ICML)}.

\bibitem[{Edunov et~al.(2018)Edunov, Ott, Auli, and Grangier}]{Edunov2018}
Sergey Edunov, Myle Ott, Michael Auli, and David Grangier. 2018.
\newblock Understanding back-translation at scale.
\newblock In \emph{Proceedings of the 2018 Conference on Empirical Methods in
  Natural Language Processing}, pages 489--500. Association for Computational
  Linguistics.

\bibitem[{Ferreira et~al.(2022)Ferreira, Thayaparan, Valentino, Rozanova, and
  Freitas}]{ferreira-etal-2022-integer}
Deborah Ferreira, Mokanarangan Thayaparan, Marco Valentino, Julia Rozanova, and
  Andre Freitas. 2022.
\newblock \href {https://doi.org/10.18653/v1/2022.findings-acl.76} {To be or
  not to be an integer? encoding variables for mathematical text}.
\newblock In \emph{Findings of the Association for Computational Linguistics:
  ACL 2022}, pages 938--948, Dublin, Ireland. Association for Computational
  Linguistics.

\bibitem[{Gehring et~al.(2017)Gehring, Auli, Grangier, Yarats, and
  Dauphin}]{Gehring2017}
Jonas Gehring, Michael Auli, David Grangier, Denis Yarats, and Yann~N. Dauphin.
  2017.
\newblock {Convolutional Sequence to Sequence Learning}.
\newblock In \emph{International Conference on Machine Learning (ICML)}.

\bibitem[{Ginev and Miller(2013)}]{Ginev2014}
Deyan Ginev and Bruce~R. Miller. 2013.
\newblock Latexml 2012 - a year of latexml.
\newblock In \emph{Intelligent Computer Mathematics (CICM)}.

\bibitem[{Goller and Kuchler(1996)}]{Goller2002}
Christoph Goller and Andreas Kuchler. 1996.
\newblock Learning task-dependent distributed representations by
  backpropagation through structure.
\newblock In \emph{Proceedings of International Conference on Neural Networks
  (ICNN'96)}.

\bibitem[{Greiner{-}Petter et~al.(2020)Greiner{-}Petter, Schubotz,
  M{\"{u}}ller, Breitinger, Cohl, Aizawa, and Gipp}]{Greiner-PetterS20}
Andr{\'{e}} Greiner{-}Petter, Moritz Schubotz, Fabian M{\"{u}}ller, Corinna
  Breitinger, Howard~S. Cohl, Akiko Aizawa, and Bela Gipp. 2020.
\newblock Discovering mathematical objects of interest - {A} study of
  mathematical notations.
\newblock In \emph{{WWW}: The Web Conference}.

\bibitem[{Greiner-Petter et~al.(2019)Greiner-Petter, Schubotz, Cohl, and
  Gipp}]{GreinerPetter2019-05-20Seman-46001}
André Greiner-Petter, Moritz Schubotz, Howard~S. Cohl, and Bela Gipp. 2019.
\newblock Semantic preserving bijective mappings for expressions involving
  special functions between computer algebra systems and document preparation
  systems.
\newblock \emph{Aslib Journal of Information Management}.

\bibitem[{Johnson et~al.(2017)Johnson, Schuster, Le, Krikun, Wu, Chen, Thorat,
  Vi{\'{e}}gas, Wattenberg, Corrado, Hughes, and Dean}]{Johnson2017}
Melvin Johnson, Mike Schuster, Quoc~V. Le, Maxim Krikun, Yonghui Wu, Zhifeng
  Chen, Nikhil Thorat, Fernanda Vi{\'{e}}gas, Martin Wattenberg, Greg Corrado,
  Macduff Hughes, and Jeffrey Dean. 2017.
\newblock {Google's Multilingual Neural Machine Translation System: Enabling
  Zero-Shot Translation}.
\newblock \emph{Transactions of the Association for Computational Linguistics}.

\bibitem[{Lample and Charton(2020)}]{lample2019deep}
Guillaume Lample and Fran{\c{c}}ois Charton. 2020.
\newblock Deep learning for symbolic mathematics.

\bibitem[{Lamport(1994)}]{lamport1994latex}
Leslie Lamport. 1994.
\newblock \emph{LATEX: a document preparation system: user's guide and
  reference manual}.
\newblock Addison-wesley.

\bibitem[{Levenshtein(1966)}]{Levenshtein1966}
V.~Levenshtein. 1966.
\newblock {Binary Codes Capable of Correcting Deletions, Insertions and
  Reversals}.
\newblock \emph{Soviet Physics Doklady}.

\bibitem[{Li et~al.(2022)Li, Zhang, Yan, Zhou, Li, Liu, and
  Cao}]{li-etal-2022-seeking}
Zhongli Li, Wenxuan Zhang, Chao Yan, Qingyu Zhou, Chao Li, Hongzhi Liu, and
  Yunbo Cao. 2022.
\newblock \href {https://doi.org/10.18653/v1/2022.findings-acl.195} {Seeking
  patterns, not just memorizing procedures: Contrastive learning for solving
  math word problems}.
\newblock In \emph{Findings of the Association for Computational Linguistics:
  ACL 2022}, pages 2486--2496, Dublin, Ireland. Association for Computational
  Linguistics.

\bibitem[{Matsuzaki et~al.(2017)Matsuzaki, Ito, Iwane, Anai, and
  Arai}]{DBLP:conf/acl/MatsuzakiIIAA17}
Takuya Matsuzaki, Takumi Ito, Hidenao Iwane, Hirokazu Anai, and Noriko~H. Arai.
  2017.
\newblock \href {https://doi.org/10.18653/v1/P17-1195} {Semantic parsing of
  pre-university math problems}.
\newblock In \emph{Proceedings of the 55th Annual Meeting of the Association
  for Computational Linguistics, {ACL} 2017, Vancouver, Canada, July 30 -
  August 4, Volume 1: Long Papers}, pages 2131--2141. Association for
  Computational Linguistics.

\bibitem[{Miller and Youssef(2003)}]{Miller2003}
Bruce Miller and Abdou Youssef. 2003.
\newblock {Technical Aspects of the Digital Library of Mathematical Functions}.
\newblock \emph{Annals of Mathematics and Artificial Intelligence}.

\bibitem[{Ott et~al.(2019)Ott, Edunov, Baevski, Fan, Gross, Ng, Grangier, and
  Auli}]{Ott2019}
Myle Ott, Sergey Edunov, Alexei Baevski, Angela Fan, Sam Gross, Nathan Ng,
  David Grangier, and Michael Auli. 2019.
\newblock fairseq: A fast, extensible toolkit for sequence modeling.
\newblock In \emph{Proceedings of the 2019 Conference of the North {A}merican
  Chapter of the Association for Computational Linguistics (Demonstrations)}.

\bibitem[{Ott et~al.(2018)Ott, Edunov, Grangier, and Auli}]{Ott2018}
Myle Ott, Sergey Edunov, David Grangier, and Michael Auli. 2018.
\newblock Scaling neural machine translation.
\newblock In \emph{Proceedings of the Third Conference on Machine Translation:
  Research Papers}. Association for Computational Linguistics.

\bibitem[{Papineni et~al.(2002)Papineni, Roukos, Ward, and
  Zhu}]{Papineni2002-bleu}
Kishore Papineni, Salim Roukos, Todd Ward, and Wei-Jing Zhu. 2002.
\newblock {BLEU: a Method for Automatic Evaluation of Machine Translation}.
\newblock In \emph{Proceedings of the 40th Annual Meeting of the Association
  for Computational Linguistics}.

\bibitem[{Paszke et~al.(2019)Paszke, Gross, Massa, Lerer, Bradbury, Chanan,
  Killeen, Lin, Gimelshein, Antiga, Desmaison, Kopf, Yang, DeVito, Raison,
  Tejani, Chilamkurthy, Steiner, Fang, Bai, and Chintala}]{2019-PyTorch}
Adam Paszke, Sam Gross, Francisco Massa, Adam Lerer, James Bradbury, Gregory
  Chanan, Trevor Killeen, Zeming Lin, Natalia Gimelshein, Luca Antiga, Alban
  Desmaison, Andreas Kopf, Edward Yang, Zachary DeVito, Martin Raison, Alykhan
  Tejani, Sasank Chilamkurthy, Benoit Steiner, Lu~Fang, Junjie Bai, and Soumith
  Chintala. 2019.
\newblock Pytorch: An imperative style, high-performance deep learning library.
\newblock In \emph{{Proc.~Neural Information Processing Systems (NeurIPS)}}.

\bibitem[{Patel et~al.(2021)Patel, Bhattamishra, and Goyal}]{patel2021nlp}
Arkil Patel, Satwik Bhattamishra, and Navin Goyal. 2021.
\newblock Are nlp models really able to solve simple math word problems?
\newblock \emph{NAACL 2021}.

\bibitem[{Pochhammer(1870)}]{Pochhammer1870}
Leo Pochhammer. 1870.
\newblock {Ueber hypergeometrische Functionen nter Ordnung.}
\newblock \emph{Journal f{\"{u}}r die reine und angewandte Mathematik (Crelles
  Journal)}.

\bibitem[{Schubotz et~al.(2018)Schubotz, Greiner-Petter, Scharpf, Meuschke,
  Cohl, and Gipp}]{SchubotzGSMCG18}
Moritz Schubotz, Andr{\'{e}} Greiner-Petter, Philipp Scharpf, Norman Meuschke,
  Howard~S Cohl, and Bela Gipp. 2018.
\newblock {Improving the Representation and Conversion of Mathematical Formulae
  by Considering their Textual Context}.
\newblock In \emph{{ACM/IEEE} Joint Conference on Digital Libraries (JCDL)}.

\bibitem[{Sutskever et~al.(2014)Sutskever, Vinyals, and
  Le}]{Sutskever2014Seq2Seq}
Ilya Sutskever, Oriol Vinyals, and Quoc~V Le. 2014.
\newblock Sequence to sequence learning with neural networks.

\bibitem[{Trinh et~al.(2018)Trinh, Dai, Luong, and Le}]{trinh2018learning}
Trieu~H Trinh, Andrew~M Dai, Minh-Thang Luong, and Quoc~V Le. 2018.
\newblock Learning longer-term dependencies in rnns with auxiliary losses.
\newblock In \emph{International Conference on Machine Learning (ICML)}.

\bibitem[{Vaswani et~al.(2017)Vaswani, Shazeer, Parmar, Uszkoreit, Jones,
  Gomez, Kaiser, and Polosukhin}]{Vaswani2017}
Ashish Vaswani, Noam Shazeer, Niki Parmar, Jakob Uszkoreit, Llion Jones,
  Aidan~N Gomez, {\L}ukasz Kaiser, and Illia Polosukhin. 2017.
\newblock Attention is all you need.
\newblock In \emph{{Proc.~Neural Information Processing Systems (NeurIPS)}}.

\bibitem[{Wang et~al.(2019)Wang, Sun, and Wang}]{Wang2019}
Jian Wang, Yunchuan Sun, and Shenling Wang. 2019.
\newblock Image to latex with densenet encoder and joint attention.

\bibitem[{Wang et~al.(2018)Wang, Kaliszyk, and Urban}]{Wang2018Mizar}
Qingxiang Wang, Cezary Kaliszyk, and Josef Urban. 2018.
\newblock {First Experiments with Neural Translation of Informal to Formal
  Mathematics}.

\bibitem[{Wu et~al.(2019)Wu, Fan, Baevski, Dauphin, and
  Auli}]{Wu2019PayLessAttention}
Felix Wu, Angela Fan, Alexei Baevski, Yann~N. Dauphin, and Michael Auli. 2019.
\newblock {Pay Less Attention with Lightweight and Dynamic Convolutions}.
\newblock In \emph{International Conference on Learning Representations
  (ICLR)}.

\bibitem[{Wu et~al.(2021)Wu, Zhang, Wei, and Huang}]{DBLP:conf/acl/WuZWH20}
Qinzhuo Wu, Qi~Zhang, Zhongyu Wei, and Xuanjing Huang. 2021.
\newblock \href {https://doi.org/10.18653/v1/2021.acl-long.455} {Math word
  problem solving with explicit numerical values}.
\newblock In \emph{Proceedings of the 59th Annual Meeting of the Association
  for Computational Linguistics and the 11th International Joint Conference on
  Natural Language Processing, {ACL/IJCNLP} 2021, (Volume 1: Long Papers),
  Virtual Event, August 1-6, 2021}, pages 5859--5869. Association for
  Computational Linguistics.

\bibitem[{Zhang et~al.(2020)Zhang, Wang, Lee, Bin, Wang, Shao, and
  Lim}]{DBLP:conf/acl/ZhangWLBWSL20}
Jipeng Zhang, Lei Wang, Roy~Ka{-}Wei Lee, Yi~Bin, Yan Wang, Jie Shao, and
  Ee{-}Peng Lim. 2020.
\newblock \href {https://doi.org/10.18653/v1/2020.acl-main.362} {Graph-to-tree
  learning for solving math word problems}.
\newblock In \emph{Proceedings of the 58th Annual Meeting of the Association
  for Computational Linguistics, {ACL} 2020, Online, July 5-10, 2020}, pages
  3928--3937. Association for Computational Linguistics.

\end{thebibliography}
}

\clearpage

\appendix
\onecolumn

\setcounter{equation}{0}
\renewcommand{\theequation}{\thesection.\arabic{equation}}

\section{Implementation Details}

For the implementation of the transformer and convolutional sequence-to-sequence models, we built on the PyTorch \cite{2019-PyTorch} library fairseq \cite{Ott2019} by Facebook AI Research.
Since, as depicted in the evaluation, the perplexity does not properly reflect the accuracy, we extended fairseq by an option to measure the quality of models with an exact match accuracy and the Levenshtein distance.

We performed the computations on GPU-accelerated hardware.
For the experiments, we used a server with 8 Nvidia Tesla V100 GPUs.
Training took, depending on the setup, $1-96$ hours.

For performance, memory, and architectural reasons, for training, we only consider formulae with up to $1\,024$ tokens following the trend of current research \cite{Bowman2016,Dauphin2016}.

\subsection{Tokenizers}

For \textbf{\LaTeX{}}, we developed a tokenizer for which \LaTeX{} commands, parentheses, braces, brackets, as well as special characters are individual tokens.
Letters are considered individual tokens and are thus split into single letters.
Multi-digit numbers are considered as tokens as described in Section~\ref{sec.substitutenumbers}.
For \textbf{Mathematica}, we developed a tokenizer that considers Mathematica functions, Symbols (e.g., \texttt{\textbackslash{}[Zeta]}), parentheses, braces, brackets, as well as special characters are individual tokens.

As most strings of letters are Mathematica functions, we also consider all strings of letters as individual tokens, i.e., we do not split them into single letters.
Multi-digit numbers are considered as tokens as described in Section~\ref{sec.substitutenumbers}.
In addition, the following are exceptional tokens: \texttt{\&\&}, \texttt{==}, \texttt{<=}, \texttt{>=},  \texttt{!=}, and \texttt{/;}.

For examples, see Supplementary Material B, which contains equations tokenized by our tokenizers. 

\vspace{4em}

\section{Qualitative Analysis -- \mbox{Mathematical Functions Site}}
\label{apx.translationExamples_mfs}

This supplementary material presents 8 translation samples from \LaTeX{} to Mathematica.
All samples are randomly selected, only restricted by the editorial constraint of fitting on a single page, and restricted to have a similar amount of exact matches and erroneous cases.
Equations B.1--B.4 are erroneous translations, while Equations B.5--B.8 lead to exact matches.

Here, the \LaTeX{} formulae were generated by Mathematica's export function.

\newpage

\begin{table}[t]
    \centering
    \caption{\textbf{[1/3]} Equations including the tokenized \LaTeX{} input, the (optional) interpretation by Mathematica (Mat.), our translation (NMT), and the ground truth (GT.). (NMT+GT.) denotes that our translation is equal to the ground truth.} \label{tab.ex.mfs}
    \includegraphics[page=2,trim=3.5cm 5.cm 3.5cm 3cm, clip, width=\linewidth]{sm_tables}
\end{table}
\begin{table}[t]
    \centering
    \caption{\textbf{[2/3]} Equations including the tokenized \LaTeX{} input, the (optional) interpretation by Mathematica (Mat.), our translation (NMT), and the ground truth (GT.). (NMT+GT.) denotes that our translation is equal to the ground truth.} 
    \includegraphics[page=3,trim=3.5cm 20.cm 3.5cm 3cm, clip, width=\linewidth]{sm_tables}
    \includegraphics[page=3,trim=3.5cm 6.75cm 3.5cm 7.5cm, clip, width=\linewidth]{sm_tables}
\end{table}
\begin{table}[t]
    \centering
    \caption{\textbf{[3/3]} Equations including the tokenized \LaTeX{} input, the (optional) interpretation by Mathematica (Mat.), our translation (NMT), and the ground truth (GT.). (NMT+GT.) denotes that our translation is equal to the ground truth.}
    \includegraphics[page=3,trim=3.5cm 4.5cm 3.5cm 21.25cm, clip, width=\linewidth]{sm_tables}
    \includegraphics[page=4,trim=3.5cm 7cm 3.5cm 3cm, clip, width=\linewidth]{sm_tables}
\end{table}

\clearpage

\section{Network Ablation Studies}
\label{ablation}

Ablation studies based on the \LaTeX{}$\to$Mathematica translation model.
The concrete results for the analysis are displayed in Tables~\ref{tab:multi-state-sizes}--\ref{tab:exp-kernel-sizes}.
For the tables, let $\texttt{C}s\texttt{x}n$ denote a convolutional encoder and equal decoder with state size $s$, kernel size 3, and $n$ consecutive layers.
Let $\texttt{C}s\texttt{ks}k\texttt{x}n$ be defined according to the previous definition but with a kernel size of $k$.
Further, let \texttt{y-z} be the concatenation of three elements: \texttt{y}, a fully connected affine layer translating between the state sizes of \texttt{y} and \texttt{z}, and \texttt{z}.
Let the embedding size equal the state size of the first layer.
For accuracy, we used the exact match accuracy on the validation set of the \LaTeX{}$\to$Mathematica translation.

\vspace{2em}

\begin{table}[h]
    \centering
    \caption{
        Experiments on mixed and constant state/embedding sizes.
    }
    \footnotesize
	\begin{tabular}{p{12em} c}
	\toprule
	Model & Acc. \\
	\midrule
	\texttt{C256x8} & $86.4\%$ \\
	\texttt{C256x12} & $88.3\%$ \\
	\texttt{C512x6-C768x4-C1024x3-} & \multirow{2}{*}{$88.6\%$} \\
	\texttt{ C2048x1-C4096x1} & \\
	\texttt{C512x4-C1024x4} & $91.2\%$ \\
	\texttt{C512x6-C768x4-C1024x2} & \multirow{1}{*}{$91.6\%$} \\
	\texttt{C512x8} & $91.9\%$ \\
	\texttt{C512x4-C1024x8} & $92.3\%$ \\
	\texttt{C512x8-C1024x4} & $92.7\%$ \\
	\texttt{C512x20} & $93.0\%$ \\
	\texttt{C512x12} & $94.9\%$ \\
	\bottomrule
	\end{tabular}
    \label{tab:multi-state-sizes}
\end{table}
\vspace{1em}
\begin{table}[h]
	\centering
	\caption{
		Additional experiments (based on \texttt{C512x8}).
	}
	\footnotesize
	\begin{tabular}{l c}
		\toprule
		Modification & Acc. \\
		\midrule
		Substitute Numbers & \textbf{$95.0\%$} \\
		Single-digit tokens & $92.7\%$ \\
		Training bias towards short formulae & $94.8\%$ \\
		Input dict.~$\neq$ output dict. & \textbf{$95.0\%$} \\
		\bottomrule
	\end{tabular}
	\label{tab:exp-additional}
\end{table}
\vspace{1em}
\begin{table}[h!]
    \centering
    \caption[Further experiment results]{
        Experiments on different numbers of layers.
    }
    \footnotesize
    \begin{tabular}{c | *{6}{c}}
        \toprule
        Model   & \texttt{C512x8} & \texttt{C512x9} & \texttt{C512x10} & \texttt{C512x11}  & \texttt{C512x12} & \texttt{C512x13} \\
        Acc. & $94.3\%$        & $94.5\%$        & $94.7\%$         & \textbf{$95.1\%$} & $95.0\%$         & $94.8\%$   \\
        \bottomrule
    \end{tabular}
    \label{tab:exp-num-layers}
\end{table}
\vspace{1em}
\begin{table}[h!]
	\centering
    \caption{
    	Experiments comparing kernel sizes (including number of parameters).
    }
	\footnotesize
	\begin{tabular}{l | *{6}{c}}
		\toprule
		Model 		   & \texttt{C512ks3x8} & \texttt{C512ks5x8} & \texttt{C512ks7x8} & \texttt{C512ks5x10} & \texttt{C512ks3x11} & \texttt{C512ks5x11} \\
	 	Acc. 	   & $94.3\%$           & $95.2\%$           & $94.1\%$           & $94.4\%$            & $95.1\%$            & $95.1\%$            \\
		Num.~of param. & $32\,671\,200$     & $49\,448\,416$     & $66\,225\,632$     & $60\,995\,040$      & $43\,699\,680$      & $66\,768\,352$      \\
		\bottomrule
	\end{tabular}
\label{tab:exp-kernel-sizes}
\end{table}

\end{document}